%% file: main.tex
\documentclass[10pt,twocolumn,letterpaper]{article}

\usepackage[pagenumbers]{wacv} 

\usepackage{graphicx}
\usepackage{amsmath}
\usepackage{amssymb}
\usepackage{booktabs}
\usepackage{diagbox}
\usepackage{threeparttable}
\usepackage{pifont}
\usepackage{multirow}
\usepackage{tabularx}
\usepackage{algorithm,algpseudocode}

\newcommand{\smallsim}{\smallsym{\mathrel}{\sim}}

\makeatletter
\newcommand{\smallsym}[2]{#1{\mathpalette\make@small@sym{#2}}}
\newcommand{\make@small@sym}[2]{%
  \vcenter{\hbox{$\m@th\downgrade@style#1#2$}}%
}
\newcommand{\downgrade@style}[1]{%
  \ifx#1\displaystyle\scriptstyle\else
    \ifx#1\textstyle\scriptstyle\else
      \scriptscriptstyle
  \fi\fi
}

\usepackage[pagebackref,breaklinks,colorlinks]{hyperref}

\usepackage[capitalize]{cleveref}
\crefname{section}{Sec.}{Secs.}
\Crefname{section}{Section}{Sections}
\Crefname{table}{Table}{Tables}
\crefname{table}{Tab.}{Tabs.}

\def\wacvPaperID{283} 
\def\confName{WACV}
\def\confYear{2025}

\begin{document}

\title{ReMix: Training Generalized Person Re-identification on a Mixture of Data}

\author{
    Timur Mamedov$^{1,2}$ \quad Anton Konushin$^{3,2}$ \quad Vadim Konushin$^{1}$ \\
    $^1$Tevian, Moscow, Russia \quad $^2$Lomonosov Moscow State University \quad $^3$AIRI, Moscow, Russia \\
    {\tt\small me@timmzak.com} \quad {\tt\small konushin@airi.net} \quad {\tt\small vadim@tevian.ai}
}
\maketitle

\input{sec/0_abstract}

\input{sec/1_intro}
\input{sec/2_related}
\input{sec/3_method}
\input{sec/4_experiments}
\input{sec/5_conclusion}

{\small
\bibliographystyle{ieee_fullname}
\bibliography{main}
}

\input{sec/supplementary}

\end{document}

%% file: sec/0_abstract.tex
\begin{abstract}
   Modern person re-identification (Re-ID) methods have a weak generalization ability and experience a major accuracy drop when capturing environments change. This is because existing multi-camera Re-ID datasets are limited in size and diversity, since such data is difficult to obtain. At the same time, enormous volumes of unlabeled single-camera records are available. Such data can be easily collected, and therefore, it is more diverse. Currently, single-camera data is used only for self-supervised pre-training of Re-ID methods. However, the diversity of single-camera data is suppressed by fine-tuning on limited multi-camera data after pre-training. In this paper, we propose ReMix, a generalized Re-ID method jointly trained on a mixture of limited labeled multi-camera and large unlabeled single-camera data. Effective training of our method is achieved through a novel data sampling strategy and new loss functions that are adapted for joint use with both types of data. Experiments show that ReMix has a high generalization ability and outperforms state-of-the-art methods in generalizable person Re-ID. To the best of our knowledge, this is the first work that explores joint training on a mixture of multi-camera and single-camera data in person Re-ID.
\end{abstract}

%% file: sec/1_intro.tex
\section{Introduction}
\label{sec:intro}

\begin{figure}[t]
    \centering
    \includegraphics[width=0.94\linewidth]{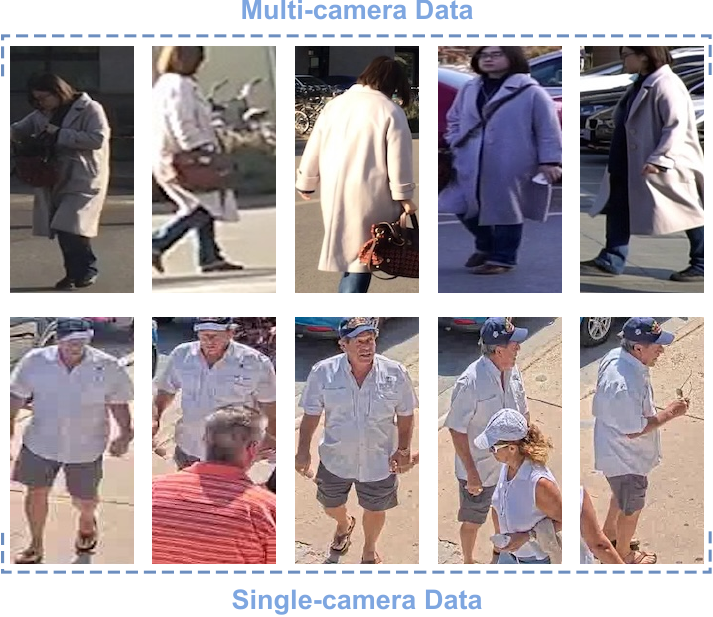}
    \caption{Examples of multi-camera and single-camera data. As we can see, multi-camera data is much more complex in terms of Re-ID: background, lighting, capturing angle, etc., may differ significantly for one person in multi-camera data. In contrast, images of the same person are less complex in single-camera data.}
    \label{fig:datasets-examples}
\end{figure}

Person re-identification (Re-ID) is the task of recognizing the same person in images taken by different cameras at different times. This task naturally arises in video surveillance and security systems, where it is necessary to track people across multiple cameras. The urgent need for robust and accurate Re-ID has stimulated scientific research over the years. However, modern Re-ID methods still have a weak generalization ability and experience a significant performance drop when capturing environments change, which limits their applicability in real-world scenarios.

\begin{table}[t]
  \centering
  \begin{tabular*}{0.47\textwidth}{l@{\hspace{5pt}}|cccc@{\hspace{7pt}}}
    \hline
    Dataset & \#images & \#IDs & \#scenes \rule{0pt}{2.3ex}\rule[-0.9ex]{0pt}{0pt}\\
    \hline
    CUHK03-NP \cite{li2014deepreid} & 14,096 & 1,467 & 2 \rule{0pt}{2.3ex}\\
    Market-1501 \cite{zheng2015scalable} & 32,668 & 1,501 & 6 \rule{0pt}{2.3ex}\\
    DukeMTMC-reID \cite{ristani2016performance} & 36,411 & 1,812 & 8 \rule{0pt}{2.3ex}\\
    MSMT17 \cite{wei2018person} & 126,441 & 4,101 & 15 \rule{0pt}{2.3ex}\\
    \hline
    LUPerson \cite{fu2021unsupervised} & $>$4M & $>$200K & 46,260 \rule{0pt}{2.3ex}\\
    \hline
  \end{tabular*}
  \caption{Comparison between existing well-known multi-camera Re-ID datasets and the single-camera LUPerson dataset. As we can see, single-camera data is more voluminous and diverse.}
  \label{tab:datasets-comparison}
\end{table}

The main reasons for the weak generalization ability of modern methods are the small amount of training data and the low diversity of capturing environments in this data. In person Re-ID, the same person may appear across multiple cameras from different angles (multi-camera data), and such data is difficult to collect and label. Due to these difficulties, each of the existing Re-ID datasets is captured from a single location. In contrast, collecting images of people from one camera (single-camera data) is much easier; for example, these images can be automatically extracted from YouTube videos \cite{fu2021unsupervised}, featuring numerous diverse identities in distinct locations and a high diversity of capturing environments (\cref{tab:datasets-comparison}).

However, single-camera data is much simpler than multi-camera data in terms of the person Re-ID task: in single-camera data, the same person can appear on only one camera and from only one angle (\cref{fig:datasets-examples}). Directly adding such simple data to the training process degrades the quality of Re-ID. Therefore, single-camera data is currently used only for self-supervised pre-training \cite{fu2021unsupervised, mamedov2023approaches}. However, we hypothesize that this approach has a limited effect on improving the generalization ability of Re-ID methods because subsequent fine-tuning for the final task is performed on relatively small and non-diverse multi-camera data.

In this paper, we propose ReMix, a generalized Re-ID method jointly trained on a mixture of limited labeled multi-camera and large unlabeled single-camera data. ReMix achieves better generalization by training on diverse single-camera data, as confirmed by our experiments. We also experimentally validate our hypothesis regarding the limitations of self-supervised pre-training and show that our joint training on two types of data overcomes them. In our ReMix method, we propose:
\begin{itemize}
    \setlength\itemsep{-0.384em}
    \item A novel data sampling strategy that allows for efficiently obtaining pseudo labels for large unlabeled single-camera data and for composing mini-batches from a mixture of images from labeled multi-camera and unlabeled single-camera datasets.
    \item A new Instance, Augmentation, and Centroids loss functions adapted for joint use with two types of data, making it possible to train ReMix. For example, the Instance and Centroids losses consider the different complexities of multi-camera and single-camera data, allowing for more efficient training of our method.
    \item Using self-supervised pre-training in combination with the proposed joint training procedure to improve pseudo labeling and the generalization ability of the algorithm.
\end{itemize}
Our experiments show that ReMix outperforms state-of-the-art methods in the cross-dataset and multi-source cross-dataset scenarios (when trained and tested on different datasets). To the best of our knowledge, this is the first work that explores joint training on a mixture of multi-camera and single-camera data in the person Re-ID task.

%% file: sec/2_related.tex
\section{Related Work}
\label{sec:related}

\subsection{Person Re-identification}

Rapid progress in solving the person re-identification task over the past few years has been associated with the emergence of CNNs. Some Re-ID approaches used the entire image to extract features \cite{zheng2017person, shen2018person, ni2021flipreid}. Other methods divided the image of a person into parts, extracted features for each part, and aggregated them to obtain full-image features \cite{wang2018learning, sun2018beyond, suh2018part}. Recently, transformer-based Re-ID methods have emerged \cite{he2021transreid, tan2022dynamic, zhang2023pha, li2023clip}, which also improve the quality of solving the problem.

Recent Re-ID methods perform well in the standard scenario, but their quality is significantly reduced when applied to datasets that differ from those used during training (when capturing environments change). In this paper, we explore the problem of weak generalization ability of existing Re-ID methods and show that it can be improved by properly using a mixture of two types of training data --- multi-camera and single-camera.

\subsection{Generalizable Person Re-identification}
Generalizable person re-identification aims to learn a robust model that performs well across various datasets. To achieve this goal, improved normalizations adapted to generalizable person Re-ID were proposed in \cite{jin2020style, choi2021meta, jiao2022dynamically}. A new residual block, consisting of multiple convolutional streams, each detecting features at a specific scale, was proposed in \cite{zhou2019omni} to create a specialized neural network architecture adapted to the person Re-ID task. In \cite{zhou2021learning}, the ideas from \cite{zhou2019omni} were continued, and an updated architecture with normalization layers was proposed to improve the generalization ability of the algorithm. Transformer-based models were also used to solve the problem under consideration: in \cite{ni2023part} it was shown that local parts of images are less susceptible to domain gap, making it more effective to compare two images by their local parts in addition to global visual information during training. In \cite{liao2022graph}, a new effective method for composing mini-batches during training was suggested, which improved the generalization ability of the algorithm.

As we can see, in most existing approaches, improving the generalization ability of the Re-ID algorithm has been achieved through the use of complex architectures. In contrast, in this paper, we prove that generalization can be achieved by properly training an efficient model using a variety of data, which is important in practice.

\begin{figure*}[th]
    \centering
    \includegraphics[width=0.98\linewidth]{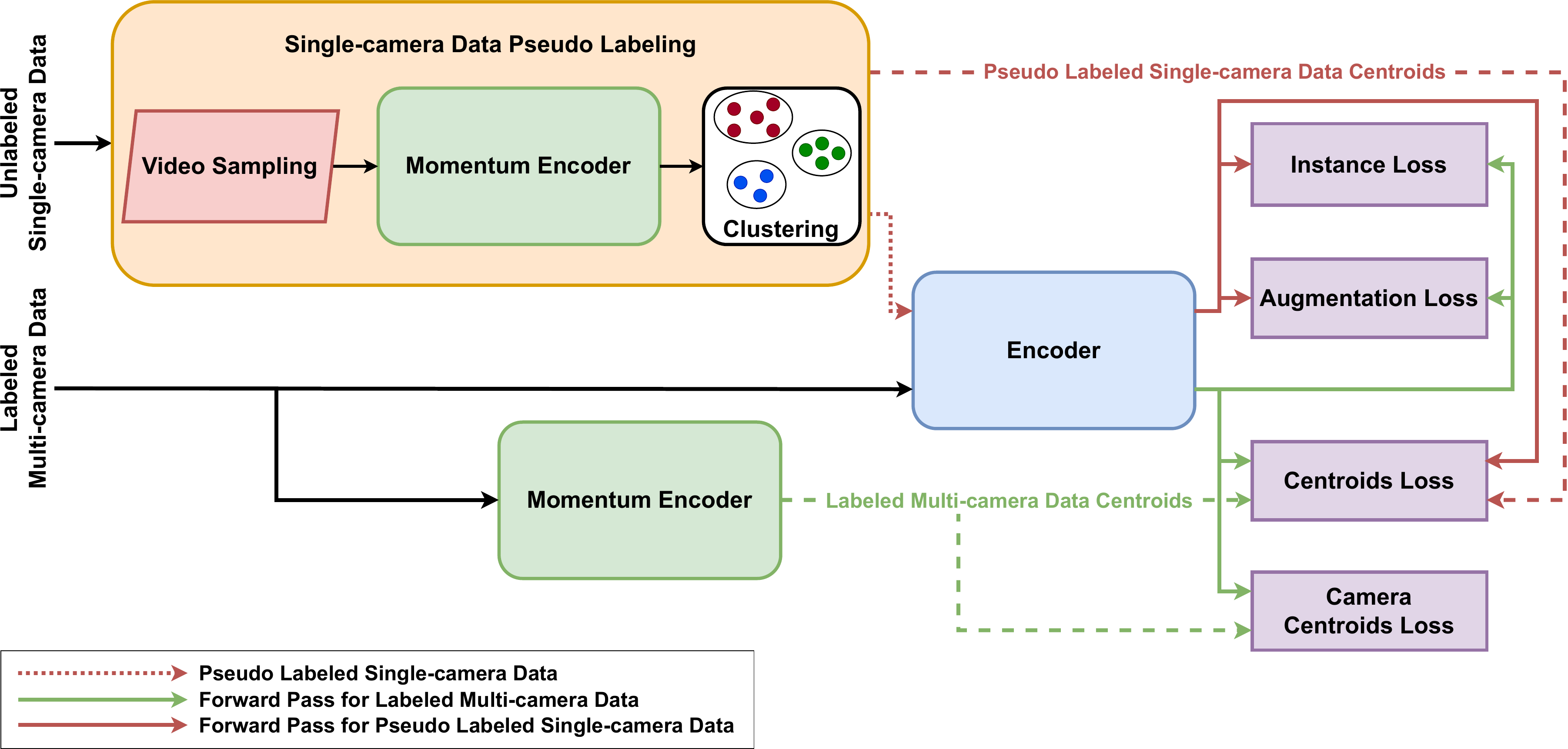}
    \caption{Scheme of ReMix. At the beginning of each epoch, all images from the person Re-ID dataset (multi-camera data) pass through the momentum encoder to obtain centroids for each identity (bottom part of the scheme). Simultaneously, videos are randomly sampled from the unlabeled single-camera dataset, and images from the selected videos are clustered using embeddings from the momentum encoder and pseudo labeled (top part of the scheme). After that, labeled multi-camera and pseudo labeled single-camera data are fed to the encoder as input. To train the encoder, the following new loss functions are used: the Instance Loss, the Augmentation Loss, and the Centroids Loss are calculated for both types of data, whereas the Camera Centroids Loss is calculated only for multi-camera data.}
    \label{fig:method-scheme}
\end{figure*}

\subsection{Self-supervised Pre-training}

Self-supervised pre-training is an approach for training neural networks using unlabeled data to learn high-quality primary features. Such pre-training is usually performed by defining relatively simple tasks that allow training data to be generated on the fly, for example: context prediction \cite{he2022masked}, solving a puzzle \cite{noroozi2016unsupervised}, predicting an image rotation angle \cite{gidaris2018unsupervised}. In \cite{chen2020simple, chen2020improved, zbontar2021barlow, dwibedi2021little}, self-supervised approaches based on contrastive learning were proposed: there, the neural network was trained to bring images of the same class closer in space and push away negative instances. Self-supervised pre-training was also used in person re-identification \cite{fu2021unsupervised, mamedov2023approaches, chen2023beyond}.

However, we suppose that this approach has a limited impact on improving the generalization ability of Re-ID methods, since subsequent fine-tuning for the final task is conducted on relatively small and non-diverse multi-camera data. In this paper, we show that the proposed joint training procedure in our ReMix method is more effective than pure self-supervised pre-training.

%% file: sec/3_method.tex
\section{Proposed Method}
\label{sec:method}

\subsection{Overview}
\label{sec:method_overview}

The scheme of ReMix is presented in \cref{fig:method-scheme}. The proposed method consists of two neural networks with identical architectures --- the encoder and the momentum encoder. The main idea of ReMix is to jointly train the Re-ID algorithm on a mixture of labeled multi-camera data for this task, and diverse unlabeled single-camera images of people. Therefore, during training, mini-batches consisting of these two types of data are used. The novel data sampling strategy is described in \cref{sec:data-sampling}.

The encoder is trained using new loss functions that are adapted for joint use with two types of data: the Instance Loss $\mathcal{L}_{ins}$ (\cref{sec:instance-loss}), the Augmentation Loss $\mathcal{L}_{aug}$ (\cref{sec:augmentation-loss}), and the Centroids Loss $\mathcal{L}_{cen}$ (\cref{sec:centroids-loss}) are calculated for both types of data, whereas the Camera Centroids Loss $\mathcal{L}_{cc}$ (\cref{sec:camera-centroids-loss}) is calculated only for multi-camera data. The general loss function in ReMix has the following form:
\begin{equation}
    \begin{split}
      \mathcal{L} &= \mathcal{L}_{ins} + \mathcal{L}_{aug} + \mathcal{L}_{cen} + \gamma \mathcal{L}_{cc}.
    \end{split}
    \label{eq:full-loss}
\end{equation}
The encoder is updated by backpropagation, and for the momentum encoder, the weights are updated using exponential moving averaging:
\begin{equation}
    \begin{split}
      \theta^{t}_{m} &= {\lambda \theta^{t-1}_{m} + (1 - \lambda) \theta^{t}_{e}},
    \end{split}
    \label{eq:momentum}
\end{equation}
where ${\theta^{t}_{e}}$ and ${\theta^{t}_{m}}$ are the weights of the encoder and the momentum encoder at iteration ${t}$, respectively; and ${\lambda}$ is the momentum coefficient.

The use of the encoder and the momentum encoder allows for more robust and noise-resistant training, which is important when using unlabeled single-camera data. During inference, only the momentum encoder is used to obtain embeddings. To train ReMix, loss functions involving centroids are applied. Therefore, to achieve training stability and frequent updating of centroids, only a portion of the images passes through the encoder in one epoch. Additionally, this approach reduces computational costs by generating pseudo labels only for a subset of single-camera data in one epoch, rather than for an entire large dataset (\cref{sec:data-sampling}). ReMix is described in more detail in the supplementary material (see \cref{alg:remix-alg}).

\subsection{Data Sampling}
\label{sec:data-sampling}

Let us formally describe the training datasets. Labeled multi-camera data (Re-ID datasets) consist of image-label-camera triples $\mathcal{D}_m = \left\{ (x_i, y_i, c_i) \right\}_{i=1}^{N_m}$, where $x_i \in \mathcal{X}$ is the image, $y_i \in \mathcal{Y}_m=\left\{ 1, 2, ..., M_m \right\}$ is the image's identity label, and $c_i \in \mathcal{C}_m=\left\{ 1, 2, ..., K_m \right\}$ is the camera ID. As for unlabeled single-camera data $\mathcal{D}_s$, it is a set of videos $\left\{ \mathcal{V}_i \right\}_{i=1}^{N_s}$, where each video $\mathcal{V}_i$ is a set of unlabeled images $\left\{ \hat{x}^i_j \right\}_{j=1}^{N_s^i}$ of people. In single-camera data, each person appears on only one video.

\vspace{2pt}\noindent\textbf{Single-camera data pseudo labeling.} Since the proposed method uses unlabeled single-camera data, pseudo labels are obtained at the beginning of each epoch. This is done according to the following algorithm: a video $\mathcal{V}_i$ is randomly sampled from the set $\mathcal{D}_s$, and images from the selected video are clustered by DBSCAN \cite{ester1996density} using embeddings from the momentum encoder and pseudo labeled. This procedure continues until pseudo labels are assigned to all images necessary for training in one epoch. As mentioned in \cref{sec:method_overview}, not all images are used for training in one epoch, so we know in advance how many images from unlabeled single-camera data should receive pseudo labels. Thus, our method iteratively obtains pseudo labels for almost all images from the large single-camera dataset. Additionally, it is worth noting that the pseudo labeling procedure uses embeddings from the momentum encoder with weights updated in the previous epoch, which leads to iterative improvements in the quality of pseudo labels. The proposed single-camera data pseudo labeling procedure is described in more detail in the supplementary material (see \cref{alg:pseudo-labeling-alg}).

\vspace{2pt}\noindent\textbf{Mini-batch composition.} In our ReMix method, we compose a mini-batch from a mixture of images from multi-camera and single-camera datasets as follows:
\begin{itemize}
    \setlength\itemsep{-0.35em}
    \item For multi-camera data, $N^m_P$ labels are randomly sampled, and for each label, $N^m_K$ corresponding images obtained from different cameras are selected.
    \item For single-camera data, $N^s_P$ pseudo labels are randomly sampled, and for each pseudo label, $N^s_K$ corresponding images are selected.
\end{itemize}
Thus, the mini-batch has a size of ${N^m_P \times N^m_K + N^s_P \times N^s_K}$ images.

\subsection{Loss Functions}

\subsubsection{The Instance Loss}
\label{sec:instance-loss}

\hspace*{10pt}The main idea of the proposed Instance Loss is to bring the anchor closer to all positive instances and push it away from all negative instances in a mini-batch. Thus, the Instance Loss forces the neural network to learn a more general solution.

Let us define $\hat{\mathcal{Y}}_{m+s} = \hat{\mathcal{Y}}_{m} \cup \hat{\mathcal{Y}}_{s}$ as the set of all labels for multi-camera data and pseudo labels for single-camera data in a mini-batch. $\hat{y}_i \in \hat{\mathcal{Y}}_{m+s}$ is either a label or pseudo label corresponding to the $i$-th image in a mini-batch. $B_m = N^m_P \times N^m_K$ is the number of images from multi-camera data in a mini-batch. And $B_s= N^s_P \times N^s_K$ is the number of images from single-camera data in a mini-batch. Then the Instance Loss is defined as follows:
\begin{equation}
    \begin{split}
      \mathcal{L}_{ins} &= \frac{1}{B_m + B_s} \Biggl( \underbrace{\sum_{i=1}^{B_m} \mathcal{L}_{ins_{m}}^i}_{\textit{multi-camera}} + \underbrace{\sum_{i=B_m + 1}^{B_m + B_s} \mathcal{L}_{ins_{s}}^i}_{\textit{single-camera}} \Biggr),
    \end{split}
    \label{eq:instance-loss}
\end{equation}
\begin{equation}
    \begin{split}
      \mathcal{L}_{ins_{m}}^i &= \frac{-1}{N^m_K} \sum_{j:\hat{y}_i = \hat{y}_j} \log{\frac{\exp(\left\langle f_i \cdot m_j \right\rangle / \tau_{ins_m})}{\sum\nolimits_{k=1}^{N_m + 1} \exp(\left\langle f_i \cdot m_k \right\rangle / \tau_{ins_m})}},
    \end{split}
    \label{eq:instance-loss-part1}
\end{equation}
\begin{equation}
    \begin{split}
      \mathcal{L}_{ins_{s}}^i &= \frac{-1}{N^s_K} \sum_{j:\hat{y}_i = \hat{y}_j} \log{\frac{\exp(\left\langle f_i \cdot m_j \right\rangle / \tau_{ins_s})}{\sum\nolimits_{k=1}^{N_s + 1} \exp(\left\langle f_i \cdot m_k \right\rangle / \tau_{ins_s})}},
    \end{split}
    \label{eq:instance-loss-part2}
\end{equation}
where $f_i$, $m_i$ are embeddings from the encoder and the momentum encoder for the anchor $i$-th image in a mini-batch, respectively; $N_m$ and $N_s$ are the numbers of negative instances for the anchor (for multi-camera and single-camera data, respectively); and ${\left\langle \cdot \right\rangle}$ denotes cosine similarity. Since multi-camera and single-camera data have different complexities in terms of person Re-ID, we balance them by using temperature parameters in the Instance Loss: $\tau_{ins_m}$ for multi-camera data and $\tau_{ins_s}$ for single-camera data.

\subsubsection{The Augmentation Loss}
\label{sec:augmentation-loss}

\hspace*{10pt}The distribution of inter-instance similarities produced by the algorithm can change under the influence of augmentations. After augmentations, an anchor image from the perspective of the neural network may become less similar to its positive pair, but at the same time, similarity to negative instances increases. Thus, current methods may be unstable to image changes and noise that may occur in practice.

To address this problem, we propose the new Augmentation Loss, which brings the augmented version of the image closer to its original and pushes it away from instances belonging to other identities in a mini-batch:
\begin{equation}
    \begin{split}
      \mathcal{L}_{aug} &= \mathbb{E} \left[ -\log{\frac{\exp(\left\langle f^i_{aug} \cdot m^i_{A} \right\rangle / \tau_{aug})}{\sum\nolimits_{j=1}^{N+1} \exp(\left\langle f^i_{aug} \cdot m_j \right\rangle / \tau_{aug})}} \right],
    \end{split}
    \label{eq:augmentation-loss}
\end{equation}
where $f^i_{aug}$ is the embedding from the encoder for the augmented $i$-th image in a mini-batch; $m^i_{A}$ is the embedding from the momentum encoder for the original $i$-th image in a mini-batch; and $N$ is the number of negative instances. It is important to note that in the Augmentation Loss, embeddings for the original images are obtained from the momentum encoder, as the momentum encoder is more stable.

\subsubsection{The Centroids Loss}
\label{sec:centroids-loss}

\hspace*{10pt}Let us define the concept of a centroid for a label or pseudo label $\hat{y}_i \in \hat{\mathcal{Y}}_{m+s}$ as follows:
\begin{equation}
    \begin{split}
      p_{\hat{y}_i} &= \frac{1}{|M_{\hat{y}_i}|} \sum_{m \in M_{\hat{y}_i}} m,
    \end{split}
\end{equation}
where $M_{\hat{y}_i}$ is the set of embeddings from the momentum encoder corresponding to the label or pseudo label $\hat{y}_i$, and ${m}$ is an embedding from this set.

Then the new Centroids Loss can be defined as:
\begin{equation}
\begin{aligned}
    \mathcal{L}_{cen} = \frac{1}{B_m + B_s} \Biggl( &\overbrace{\sum_{i=1}^{B_m} \mathcal{L}_{cen}^i(\tau_{cen_m})}^{\textit{multi-camera}} \\
    &+ \underbrace{\sum_{i=B_m + 1}^{B_m + B_s} \mathcal{L}_{cen}^i(\tau_{cen_s})}_{\textit{single-camera}} \Biggr),
    \end{aligned}
    \label{eq:centroids-loss}
\end{equation}
\begin{equation}
    \begin{split}
      \mathcal{L}_{cen}^i(\tau) &= -\log{\frac{\exp(f_{\hat{y}_i} \cdot p_{\hat{y}_i} / \tau)}{\sum\nolimits_{j=1}^{|\hat{Y}_{m+s}|} \exp(f_{\hat{y}_i} \cdot p_{\hat{y}_j} / \tau)}},
    \end{split}
\end{equation}
where $f_{\hat{y}_i}$ is the embedding from the encoder for the image with the label or pseudo label $\hat{y}_i$. Thus, this loss function brings instances closer to their corresponding centroids and pushes them away from other centroids. Like the Instance Loss, this loss function uses different temperature parameters for multi-camera and single-camera data.

\subsubsection{The Camera Centroids Loss}
\label{sec:camera-centroids-loss}

\hspace*{10pt}Since the same person could be captured by different cameras in multi-camera data, it is useful to apply information about cameras for better feature generation. In our ReMix method, we use the Camera Centroids Loss \cite{wang2021camera}. This loss function brings instances closer to the centroids of instances with the same label, but captured by different cameras. Thus, the intra-class variance caused by stylistic differences between cameras is reduced.

%% file: sec/4_experiments.tex
\section{Experiments}
\label{sec:experiments}

\subsection{Datasets and Evaluation Metrics}

\vspace{2pt}\noindent\textbf{Multi-camera datasets.} We employ well-known datasets CUHK03-NP \cite{li2014deepreid}, Market-1501 \cite{zheng2015scalable}, DukeMTMC-reID \cite{ristani2016performance}, and MSMT17 \cite{wei2018person} as multi-camera data for evaluating our proposed method. The CUHK03-NP dataset consists of 14,096 images of 1,467 identities captured by two cameras. Market-1501 was gathered from six cameras and consists of 12,936 images of 751 identities for training and 19,732 images of 750 identities for testing. DukeMTMC-reID contains 16,522 training images of 702 identities and 19,889 images of 702 identities for testing, all of them collected from eight cameras. MSMT17, a large-scale Re-ID dataset, consists of 32,621 training images of 1,041 identities and 93,820 testing images of 3,060 identities captured by fifteen cameras. Additionally, we use MSMT17-merged, which combines training and test parts. We also employ a subset of the synthetic RandPerson \cite{wang2020surpassing} dataset, which contains 132,145 training images of 8,000 identities, for additional experiments. It is worth noting that DukeMTMC-reID was withdrawn by its creators due to ethical concerns, but this dataset is still used to evaluate other modern Re-ID methods. Therefore, we include it in our tests for fair and objective comparison.

\vspace{2pt}\noindent\textbf{Single-camera dataset.} We use the LUPerson dataset \cite{fu2021unsupervised} as unlabeled single-camera data. This dataset consists of over 4 million images of more than 200,000 people from 46,260 distinct locations. To collect it, YouTube videos were automatically processed. As we can see, this dataset is much larger than multi-camera datasets for person Re-ID and covers a much more diverse range of capturing environments (\cref{tab:datasets-comparison}). Therefore, this kind of data is also useful for training Re-ID algorithms.

\vspace{2pt}\noindent\textbf{Metrics.} In our experiments, we use Cumulative Matching Characteristics (CMC) $Rank_1$, as well as mean Average Precision ($mAP$) to evaluate our method.

\subsection{Implementation Details}

In this paper, we use ResNet50 \cite{he2016deep} with IBN-a \cite{pan2018two} layers as the encoder and the momentum encoder. These encoders are self-supervised pre-trained on single-camera data from LUPerson using MoCo v2 \cite{chen2020improved}. Adam is used as an optimizer with a learning rate of $0.00035$, a weight decay rate of $0.0005$, and with a warm-up scheme in the first 10 epochs. As for the momentum coefficient $\lambda$ in \cref{eq:momentum}, we set $\lambda = 0.999$. ReMix is trained for 100 epochs. In our experiments, we set $N^m_P=N^s_P=8$ and  $N^m_K=N^s_K=4$, so the size of each mini-batch is 64. According to \cite{wang2021camera}, we choose $\gamma=0.5$ in \cref{eq:full-loss}. In ReMix, all images are resized to ${256 \times 128}$, random crops, horizontal flipping, Gaussian blurring, and random grayscale are also applied to them.

\subsection{Parameter Analysis}

\subsubsection{Temperature Parameters}
\label{sec:temperature-parameters-analysis}

\begin{table}[t]
  \centering
  \begin{subtable}{0.47\textwidth}
    \centering
    \begin{tabular*}{\textwidth}{@{\hspace{5pt}\extracolsep{\fill}}c|*{3}{c}@{\hspace{5pt}}}
      \hline
      \diagbox{$\tau_{ins_m}$}{$\tau_{aug}$} & $0.07$ & $0.10$ & $0.15$ \\
      \hline
      $0.07$ & $75.1/58.4$ & $75.1/58.5$ & $74.9/58.3$ \rule{0pt}{2.3ex}\\
      $0.10$ & $75.1/58.7$ & $\mathbf{75.8/58.7}$ & $75.0/58.6$ \rule{0pt}{2.3ex}\\
      $0.15$ & $75.0/58.3$ & $75.0/58.5$ & $74.6/58.2$ \rule{0pt}{2.3ex}\\
      \hline
    \end{tabular*}
    \subcaption{Analysis of values for parameters $\tau_{ins_m}$ and $\tau_{aug}$ in \cref{eq:instance-loss-part1} and \cref{eq:augmentation-loss}. In this table, the first number is $Rank_1$, and the second is $mAP$.}
    \label{tab:tau-ins-m-and-tau-aug}
  \end{subtable}
  \vspace{0.25cm}
  
  \begin{subtable}{0.47\textwidth}
    \centering
    \begin{tabular*}{\textwidth}{@{\hspace{7pt}\extracolsep{\fill}}l|*{5}{c}@{\hspace{5pt}}}
      \hline
      $\tau_{ins_s}$ & $0.07$ & $0.10$ & $0.15$ & $0.20$ & $0.25$ \rule{0pt}{2.3ex}\rule[-0.9ex]{0pt}{0pt}\\
      \hline
      $Rank_1$ & $75.7$ & $76.2$ & $76.3$ & $\mathbf{76.3}$ & $74.9$ \rule{0pt}{2.3ex}\\
      $mAP$ & $59.1$ & $59.6$ & $59.6$ & $\mathbf{59.9}$ & $59.5$ \rule{0pt}{2.3ex}\\
      \hline
    \end{tabular*}
    \subcaption{Analysis of values for parameter $\tau_{ins_s}$ in \cref{eq:instance-loss-part2}.}
    \label{tab:tau-ins-s}
  \end{subtable}
  
  \vspace{0.25cm}
  
  \begin{subtable}{0.47\textwidth}
    \centering
    \begin{tabular*}{\textwidth}{@{\hspace{7pt}\extracolsep{\fill}}l|*{4}{c}@{\hspace{5pt}}}
      \hline
      $\tau_{cen_s}$ \hspace{1cm}& $0.40$ & $0.50$ & $0.60$ & $0.65$ \rule{0pt}{2.3ex}\rule[-0.9ex]{0pt}{0pt}\\
      \hline
      $Rank_1$ & $76.3$ & $76.3$ & $\mathbf{76.9}$ & $75.8$ \rule{0pt}{2.3ex}\\
      $mAP$ & $60.0$ & $60.4$ & $\mathbf{60.7}$ & $60.4$ \rule{0pt}{2.3ex}\\
      \hline
    \end{tabular*}
    \subcaption{Analysis of values for parameter $\tau_{cen_s}$ in \cref{eq:centroids-loss}.}
    \label{tab:tau-cen-s}
  \end{subtable}
  \caption{Temperature parameters analysis. In these experiments, we train the algorithm on MSMT17-merged and single-camera data from LUPerson, and test it on DukeMTMC-reID.}
  \label{tab:combined-tables}
\end{table}

\vspace{2pt}\noindent\textbf{Multi-camera parameters analysis.} First, we analyze the quality of our ReMix method for different values of parameters $\tau_{ins_m}$ and $\tau_{aug}$ in the Instance Loss (\cref{sec:instance-loss}) and the Augmentation Loss (\cref{sec:augmentation-loss}), respectively. Single-camera data is not used in these experiments. As can be seen from \cref{tab:tau-ins-m-and-tau-aug}, the best quality of cross-dataset Re-ID can be achieved with $\tau_{ins_m}=\tau_{aug}=0.1$. According to \cite{chen2021ice}, we choose $\tau_{cen_m}=0.5$ in \cref{eq:centroids-loss}.

\vspace{2pt}\noindent\textbf{Single-camera parameters analysis.} Multi-camera and single-camera data have different complexities in terms of person Re-ID. So, in the Instance Loss (\cref{sec:instance-loss}) and the Centroids Loss (\cref{sec:centroids-loss}) we propose to use special temperature parameters for single-camera data ($\tau_{ins_s}$ and $\tau_{cen_s}$, respectively). According to \cref{tab:tau-ins-s} and \cref{tab:tau-cen-s}, the best results achieved when $\tau_{ins_s}=0.2$ and $\tau_{cen_s}=0.6$.

\vspace{2pt}\noindent\textbf{Conclusions from the analysis.} The temperature parameters $\tau_{ins_m}=0.1$ and $\tau_{cen_m}=0.5$ are selected for multi-camera data, $\tau_{ins_s}=0.2$ and $\tau_{cen_s}=0.6$ are selected for single-camera data. Higher temperature values make the probabilities closer together, which complicates training on simpler single-camera data. Accordingly, we confirm our hypothesis about the different complexities of multi-camera and single-camera data.

\subsubsection{Epoch Duration}
\label{sec:epoch-duration-analysis}

\hspace*{10pt}To achieve training stability and frequent updating of centroids, only a portion of the images is used during one epoch (\cref{sec:method_overview}). Also, this approach reduces computational costs by generating pseudo labels only for a subset of single-camera data in one epoch, rather than for an entire large dataset (\cref{sec:data-sampling}). In this paper, one epoch consists of 400 iterations. As can be seen from the experimental results presented in \cref{tab:iters-per-epoch}, this number of iterations is a trade-off between the accuracy of our method and its training time.

\begin{table}[t]
  \centering
  \begin{tabular*}{0.47\textwidth}{@{\hspace{10pt}\extracolsep{\fill}}l|*{4}{c}@{\hspace{5pt}}}
    \hline
    Iterations \hspace{1cm}& $300$ & $400$ & $600$ & $800$ \rule{0pt}{2.3ex}\rule[-0.9ex]{0pt}{0pt}\\
    \hline
    $Rank_1$ & $76.4$ & $\mathbf{77.6}$ & $77.1$ & $77.2$ \rule{0pt}{2.3ex}\\
    $mAP$ & $61.1$ & $61.6$ & $\mathbf{62.0}$ & $61.6$ \rule{0pt}{2.3ex}\\
    \hline
    Training Time* & $\smallsim$15h & $\smallsim$20h & $\smallsim$30h & $\smallsim$40h \rule{0pt}{2.3ex}\\
    \hline
  \end{tabular*}
  \begin{tablenotes}\small
      \centering
      \item[1] * Two Nvidia GTX 1080 Ti are used for training.
  \end{tablenotes}
  \caption{Comparison of different numbers of iterations in one epoch. We train the algorithm on MSMT17-merged and single-camera data from LUPerson, and test it on DukeMTMC-reID.}
  \label{tab:iters-per-epoch}
\end{table}

\subsection{Ablation Study}

\begin{table}[t]
  \centering
  \begin{tabular*}{0.47\textwidth}{@{\hspace{1.5pt}}cc|cc|cc}
    \hline
    \multicolumn{2}{@{\hspace{1pt}}l|}{Using s-cam. data} & \multicolumn{2}{c|}{Market-1501} & \multicolumn{2}{@{\hspace{3pt}}c}{DukeMTMC-reID}\\
    Pre-train & Joint & $Rank_1$ & $mAP$ & $Rank_1$ & $mAP$ \rule{0pt}{2.3ex}\rule[-0.9ex]{0pt}{0pt}\\
    \hline
    \ding{55} & \ding{55} & $78.4$ & $51.7$ & $75.8$ & $58.7$ \rule{0pt}{2.3ex}\\
    \ding{51} & \ding{55} & $81.7$ & $54.9$ & $75.1$ & $59.2$ \rule{0pt}{2.3ex}\\
    \ding{55} & \ding{51} & $81.3$ & $57.0$ & $76.9$ & $60.7$ \rule{0pt}{2.3ex}\\
    \ding{51} & \ding{51} & $\mathbf{84.0}$ & $\mathbf{61.0}$ & $\mathbf{77.6}$ & $\mathbf{61.6}$ \rule{0pt}{2.3ex}\\
    \hline
  \end{tabular*}
  \caption{Impact of using single-camera data in self-supervised pre-training and in our joint training procedure. In these experiments, we use MSMT17-merged and single-camera data from LUPerson (where applicable) for training.}
  \label{tab:ablation-overall}
\end{table}

\begin{table}[t]
  \centering
  \begin{tabular*}{0.47\textwidth}{@{\hspace{10pt}\extracolsep{\fill}}p{4.4cm}|cc@{\hspace{8pt}}}
    \hline
    Configuration & $Rank_1$ & $mAP$ \rule{0pt}{2.3ex}\rule[-0.9ex]{0pt}{0pt}\\
    \hline
    w/o single-camera data & $75.8$ & $58.7$ \rule{0pt}{2.3ex}\\
    $+$ in $\mathcal{L}_{aug}$  & $76.0$ & $59.2$ \rule{0pt}{2.3ex}\\
    $+$ in $\mathcal{L}_{ins}$  & $76.3$ & $59.9$ \rule{0pt}{2.3ex}\\
    $+$ in $\mathcal{L}_{cen}$ only as centroids & $75.4$ & $60.0$ \rule{0pt}{2.3ex}\\
    $+$ in $\mathcal{L}_{cen}$ & $\mathbf{76.9}$ & $\mathbf{60.7}$ \rule{0pt}{2.3ex}\\
    \hline
  \end{tabular*}
  \caption{Step-by-step use of single-camera data in different loss functions. Here, "in $\mathcal{L}_{cen}$ only as centroids" means that single-camera data is used only as centroids in the Centroids Loss. We train the algorithm on MSMT17-merged and single-camera data from LUPerson, and test it on DukeMTMC-reID.}
  \label{tab:ablation-single-cams}
\end{table}

\begin{table*}[t]
  \centering
  \begin{subtable}[b]{0.495\linewidth}
    \centering
    \begin{tabularx}{0.93\linewidth}{@{\hspace{7pt}}l|c|cc}
        \hline
        \multirow{2}{*}{Method} & \multirow{2}{*}{Reference} & \multicolumn{2}{c}{Market-1501}\\
        & & $Rank_1$ & $mAP$ \rule{0pt}{2.3ex}\rule[-1.0ex]{0pt}{0pt}\\
        \hline
        SNR \cite{jin2020style} & CVPR20 & $66.7$ & $33.9$ \rule{0pt}{2.3ex}\\
        MetaBIN \cite{choi2021meta} & CVPR21 & $69.2$ & $35.9$ \rule{0pt}{2.3ex}\\
        MDA \cite{ni2022meta} & CVPR22 & $70.3$ & $38.0$ \rule{0pt}{2.3ex}\\
        DTIN-Net \cite{jiao2022dynamically} & ECCV22 & $69.8$ & $37.4$ \rule{0pt}{2.3ex}\\
        ReMix (w/o s-cam.) & Ours & $68.2$ & $37.7$ \rule{0pt}{2.3ex}\\
        ReMix & Ours & $\mathbf{71.3}$ & $\mathbf{43.0}$ \rule{0pt}{2.3ex}\\
        \hline
    \end{tabularx}
    \caption{Training dataset: DukeMTMC-reID.}
  \end{subtable}
  \begin{subtable}[b]{0.495\linewidth}
    \centering
    \begin{tabularx}{0.93\linewidth}{@{\hspace{5pt}}l|c|cc@{\hspace{2pt}}}
        \hline
        \multirow{2}{*}{Method} & \multirow{2}{*}{Reference} & \multicolumn{2}{c}{DukeMTMC-reID}\\
        & & $Rank_1$ & $mAP$ \rule{0pt}{2.3ex}\rule[-1.0ex]{0pt}{0pt}\\
        \hline
        SNR \cite{jin2020style} & CVPR20 & $55.1$ & $33.6$ \rule{0pt}{2.3ex}\\
        MetaBIN \cite{choi2021meta} & CVPR21 & $55.2$ & $33.1$ \rule{0pt}{2.3ex}\\
        MDA \cite{ni2022meta} & CVPR22 & $56.7$ & $34.4$ \rule{0pt}{2.3ex}\\
        DTIN-Net \cite{jiao2022dynamically} & ECCV22 & $57.0$ & $36.1$ \rule{0pt}{2.3ex}\\
        ReMix (w/o s-cam.) & Ours & $57.1$ & $36.5$ \rule{0pt}{2.3ex}\\
        ReMix & Ours & $\mathbf{58.4}$ & $\mathbf{38.8}$ \rule{0pt}{2.3ex}\\
        \hline
    \end{tabularx}
    \caption{Training dataset: Market-1501.}
  \end{subtable}

  \vspace{0.25cm}

  \begin{subtable}[b]{\linewidth}
    \centering
    \begin{tabularx}{0.96\linewidth}{@{\hspace{5pt}}l|c|c|cc|cc|cc@{\hspace{2pt}}}
        \hline
        \multirow{2}{*}{Method} & \multirow{2}{*}{Reference} & \multirow{2}{*}{Training Dataset} & \multicolumn{2}{c|}{CUHK03-NP} & \multicolumn{2}{c|}{Market-1501} & \multicolumn{2}{c}{DukeMTMC-reID}\\
        & & & $Rank_1$ & $mAP$ & $Rank_1$ & $mAP$ & $Rank_1$ & $mAP$ \rule[-1.0ex]{0pt}{0pt}\\
        \hline
        SNR \cite{jin2020style} & CVPR20 & \multirow{7}{*}{MSMT17} & --- & --- & $70.1$ & $41.4$ & $69.2$ & $50.0$ \rule{0pt}{2.3ex}\\
        QAConv \cite{liao2020interpretable} & ECCV20 & & $25.3$ & $22.6$ & $72.6$ & $43.1$ & $69.4$ & $52.6$ \rule{0pt}{2.3ex}\\
        TransMatcher \cite{liao2021transmatcher} & NeurIPS21 & & $23.7$ & $22.5$ & $\mathbf{80.1}$ & $52.0$ & --- & --- \rule{0pt}{2.3ex}\\
        QAConv-GS \cite{liao2022graph} & CVPR22 & & $20.9$ & $20.6$ & $79.1$ & $49.5$ & $67.3$ & $49.4$ \rule{0pt}{2.3ex}\\
        PAT \cite{ni2023part} & ICCV23 & & $24.2$ & $25.1$ & $72.2$ & $47.3$ & --- & --- \rule{0pt}{2.3ex}\\
        ReMix (w/o s-cam.) & Ours & & $24.1$ & $24.5$ & $73.0$ & $42.5$ & $68.9$ & $49.2$ \rule{0pt}{2.3ex}\\
        ReMix & Ours & & $\mathbf{27.3}$ & $\mathbf{27.4}$ & $78.2$ & $\mathbf{52.4}$ & $\mathbf{71.6}$ & $\mathbf{52.8}$ \rule[-1.0ex]{0pt}{0pt}\\
        \hline
        OSNet \cite{zhou2019omni} & CVPR19 & \multirow{6}{*}{MSMT17-merged} & --- & --- & $66.5$ & $37.2$ & --- & --- \rule{0pt}{2.3ex}\\
        OSNet-AIN \cite{zhou2021learning} & TPAMI21 & & --- & --- & $70.1$ & $43.3$ & --- & --- \rule{0pt}{2.3ex}\\
        TransMatcher \cite{liao2021transmatcher} & NeurIPS21 & & $31.9$ & $30.7$ & $82.6$ & $58.4$ & --- & --- \rule{0pt}{2.3ex}\\
        QAConv-GS \cite{liao2022graph} & CVPR22 & & $27.6$ & $28.0$ & $80.6$ & $55.6$ & $71.3$ & $53.5$ \rule{0pt}{2.3ex}\\
        ReMix (w/o s-cam.) & Ours & & $34.5$ & $32.7$ & $78.4$ & $51.7$ & $75.8$ & $58.7$ \rule{0pt}{2.3ex}\\
        ReMix & Ours & & $\mathbf{37.7}$ & $\mathbf{37.2}$ & $\mathbf{84.0}$ & $\mathbf{61.0}$ & $\mathbf{77.6}$ & $\mathbf{61.6}$ \rule[-1.0ex]{0pt}{0pt}\\
        \hline
        RP Baseline \cite{wang2020surpassing} & ACMMM20 & \multirow{5}{*}{RandPerson} & $13.4$ & $10.8$ & $55.6$ & $28.8$ & --- & --- \rule{0pt}{2.3ex}\\
        CBN \cite{zhang2021unrealperson} & ECCV20 & & --- & --- & $64.7$ & $39.3$ & --- & --- \rule{0pt}{2.3ex}\\
        QAConv-GS \cite{liao2022graph} & CVPR22 & & $14.8$ & $13.4$ & $\mathbf{74.0}$ & $43.8$ & --- & --- \rule{0pt}{2.3ex}\\
        ReMix (w/o s-cam.) & Ours & & $17.1$ & $15.7$ & $71.1$ & $42.4$ & $61.2$ & $39.0$ \rule{0pt}{2.3ex}\\
        ReMix & Ours & & $\mathbf{19.3}$ & $\mathbf{18.4}$ & $72.7$ & $\mathbf{45.4}$ & $\mathbf{63.2}$ & $\mathbf{42.8}$ \rule{0pt}{2.3ex}\\
        \hline
    \end{tabularx}
    \caption{Training datasets: MSMT17, MSMT17-merged, and RandPerson.}
  \end{subtable}
  \caption{Comparison of our ReMix method with others in the cross-dataset scenario. In this comparison, we use two versions of the proposed method: without using single-camera data and with using single-camera data during training. Here, we use the LUPerson dataset as single-camera data to train ReMix.}
  \label{tab:comparison}
\end{table*}

\vspace{2pt}\noindent\textbf{Proof-of-concept.} We conduct a series of experiments to demonstrate the effectiveness of the proposed idea of joint training on multi-camera and single-camera data. The results of these experiments are presented in \cref{tab:ablation-overall}. As we can see, using single-camera data in addition to multi-camera data significantly improves the generalization ability of the algorithm and the quality of cross-dataset Re-ID. It is worth noting that the use of single-camera data most significantly affects the $mAP$ metric. That is, our method produces higher similarity values for images of the same person and lower values for different ones. This is achieved due to a more diverse training data, which is primarily obtained from large amounts of single-camera data.

Moreover, the effectiveness of our approach is demonstrated in comparison with self-supervised pre-training: the model trained using the proposed joint training procedure achieves better accuracy than the self-supervised pre-trained model. In \cref{sec:intro} we hypothesized that self-supervised pre-training has a limited effect, since subsequent fine-tuning for the final task is performed on relatively small multi-camera data. The results of our experiments validate this hypothesis. Indeed, by using our joint training procedure together with self-supervised pre-training, we can achieve the best quality. Thus, we experimentally confirm the importance of data volume at the fine-tuning stage. ReMix uses unlabeled single-camera data, and this result can also verify that self-supervised pre-training improves the quality of clustering and pseudo labeling.

\vspace{2pt}\noindent\textbf{Using single-camera data in loss functions.} In addition to experiments showing the validity of our joint training procedure, we conduct an ablation study to demonstrate the effectiveness of adapting the proposed loss functions for joint use with two types of data. In this study, we gradually add single-camera data in losses and measure the final accuracy. As we can see from \cref{tab:ablation-single-cams}, each loss function added to a combination improves the performance, and using all losses with single-camera data jointly provides the highest quality. Thus, the proposed loss functions are successfully adapted for joint use with two types of training data --- multi-camera and single-camera.

\subsection{Comparison with State-of-the-Art Methods}
\label{sec:comparison}

\begin{table*}[t]
  \centering
  \begin{tabular}{@{\hspace{5pt}}l|c|cc|cc|cc@{\hspace{2pt}}}
    \hline
    \multirow{2}{*}{Method} & \multirow{2}{*}{Reference} & \multicolumn{2}{c|}{M+D+MS $\rightarrow$ C3} & \multicolumn{2}{c|}{D+C3+MS $\rightarrow$ M} & \multicolumn{2}{c}{M+C3+MS $\rightarrow$ D}\\
    & & $Rank_1$ & $mAP$ & $Rank_1$ & $mAP$ & $Rank_1$ & $mAP$\rule{0pt}{2.3ex}\rule[-1.0ex]{0pt}{0pt}\\
    \hline
    MECL \cite{yu2021multiple} & arXiv21 & $32.1$ & $31.5$ & $80.0$ & $56.5$ & $70.0$ & $53.4$\rule{0pt}{2.3ex}\\
    M$^3$L \cite{zhao2021learning} & ICCV21 & $36.4$ & $35.2$ & $81.5$ & $59.6$ & $71.8$ & $54.5$\rule{0pt}{2.3ex}\\
    RaMoE \cite{dai2021generalizable} & CVPR21 & $36.6$ & $35.5$ & $82.0$ & $56.5$ & $73.6$ & $56.9$\rule{0pt}{2.3ex}\\
    MetaBIN \cite{choi2021meta} & CVPR21 & $38.1$ & $37.5$ & $83.2$ & $61.2$ & $71.3$ & $54.9$\rule{0pt}{2.3ex}\\
    MixNorm \cite{qi2022novel} & TMM22 & $29.6$ & $29.0$ & $78.9$ & $51.4$ & $70.8$ & $49.9$\rule{0pt}{2.3ex}\\
    META \cite{xu2021meta} & ECCV22 & $46.0$ & $45.9$ & $85.3$ & $65.7$ & $76.9$ & $59.9$\rule{0pt}{2.3ex}\\
    IL \cite{tan2023style} & TMM23 & $40.9$ & $38.3$ & $86.2$ & $65.8$ & $75.4$ & $57.1$\rule{0pt}{2.3ex}\\
    ReMix & Ours & $\mathbf{47.6}$ & $\mathbf{46.5}$ & $\mathbf{87.8}$ & $\mathbf{70.5}$ & $\mathbf{79.0}$ & $\mathbf{63.3}$\rule[-1.0ex]{0pt}{0pt}\\
    \hline
  \end{tabular}
  \caption{Comparison of our ReMix method with others in the multi-source cross-dataset scenario. Here, we use the LUPerson dataset as single-camera data to train ReMix. In this table, C3 is CUHK03-NP, M is Market-1501, D is DukeMTMC-reID, and MS is MSMT17.}
  \label{tab:comparison-multi}
\end{table*}

\begin{figure}[t]
    \centering
    \includegraphics[width=0.96\linewidth]{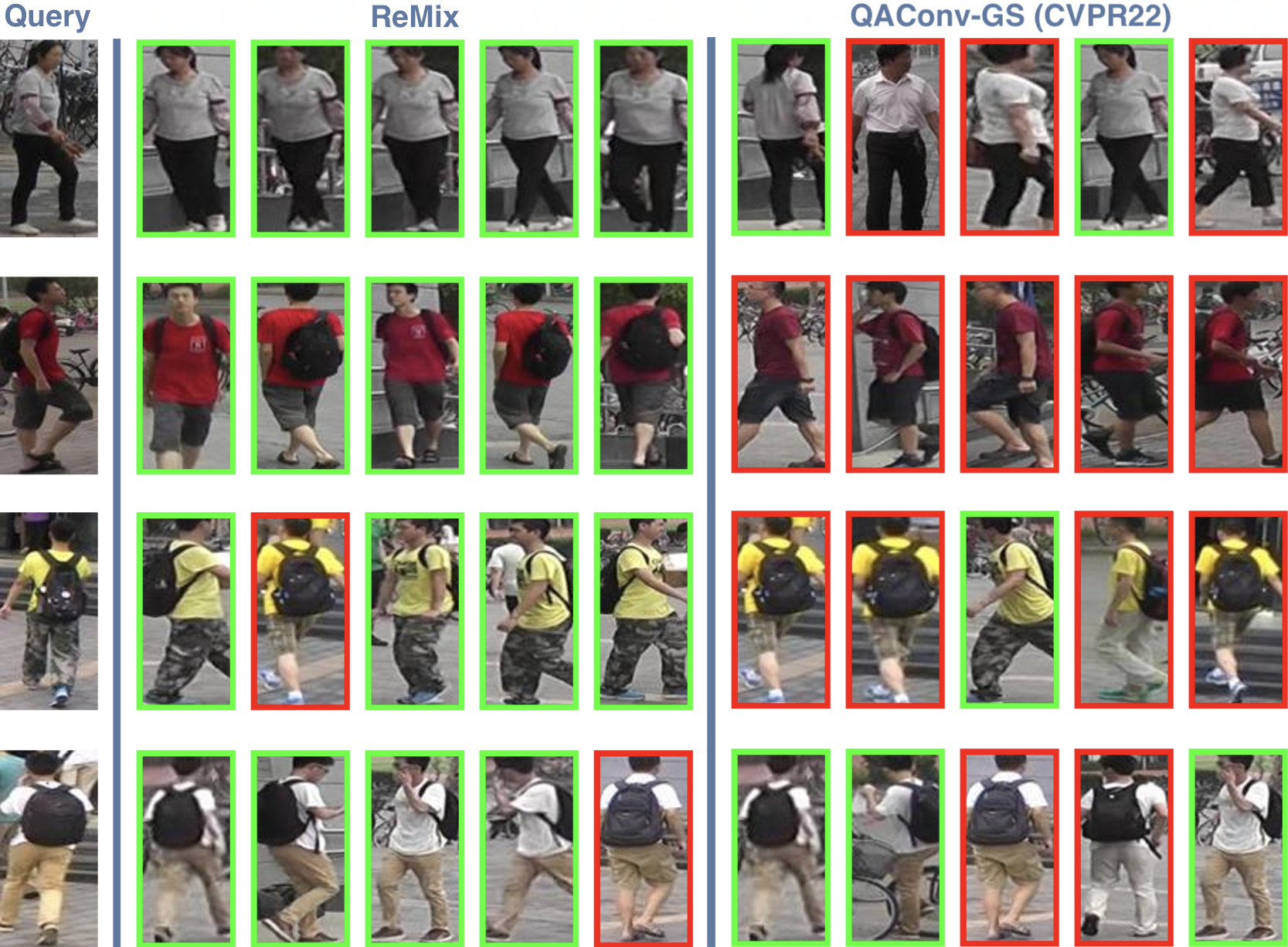}
    \caption{Comparison of TOP-5 retrieved images on the Market-1501 dataset between ReMix and QAConv-GS \cite{liao2022graph}. Green boxes denote correct results, while red boxes denote incorrect results.}
    \label{fig:ranking-visualization}
\end{figure}

We compare our ReMix method with other state-of-the-art Re-ID approaches using two test protocols: the cross-dataset and multi-source cross-dataset scenarios. According to the first protocol, we train the algorithm on one multi-camera dataset and test it on another multi-camera dataset. In the multi-source cross-dataset scenario, we train the algorithm on several multi-camera datasets and test it on another multi-camera dataset. Thus, we evaluate the generalization ability of our method in comparison to other existing state-of-the-art Re-ID approaches. Also, we illustrate several complex examples in \cref{fig:ranking-visualization}, where ReMix manages to notice important visual cues.

\vspace{2pt}\noindent\textbf{The cross-dataset scenario.} As can be seen from \cref{tab:comparison}, the proposed method demonstrates a high generalization ability and outperforms others in the cross-dataset scenario. In our ReMix method, the momentum encoder is trained to obtain embeddings for each query and gallery image, after which they are compared using cosine similarity. QAConv \cite{liao2020interpretable}, TransMatcher \cite{liao2021transmatcher}, and QAConv-GS \cite{liao2022graph}, which are among the most accurate methods in cross-dataset person Re-ID, use more complex architectures: in addition to the encoder, a separate neural network is used. This network compares features between the query and gallery images and predicts the probability that they belong to the same person. PAT \cite{ni2023part} uses a transformer-based model, which is more computationally complex compared to ResNet50 with IBN-a layers in ReMix. Thus, most existing state-of-the-art approaches improve generalization ability by using complex architectures. In contrast, the high performance of our method is achieved through the training strategy that does not affect the computational complexity, so our method can seamlessly replace other methods used in real-world applications. It is also worth noting that in the comparison in \cref{tab:comparison}, some methods use larger input images. In the supplementary material, we show that the accuracy of ReMix increases with the size of the input image (see \cref{sec:input-images-size}).

\vspace{2pt}\noindent\textbf{The multi-source cross-dataset scenario.} The comparison presented in \cref{tab:comparison-multi} shows the effectiveness of our joint training procedure, even when using several multi-camera datasets and one single-camera dataset during training. This further proves the consistency and flexibility of ReMix.

%% file: sec/5_conclusion.tex
\section{Conclusion}
\label{sec:conclusion}

In this paper, we proposed ReMix, a novel person Re-ID method that achieves generalization by jointly using limited labeled multi-camera and large unlabeled single-camera data for training. To the best of our knowledge, this is the first work that explores joint training on a mixture of multi-camera and single-camera data in person Re-ID. To provide effective training, we developed a novel data sampling strategy and new loss functions adapted for joint use with these two types of data. Through experiments, we showed that our method has a high generalization ability and outperforms state-of-the-art methods in the cross-dataset and multi-source cross-dataset scenarios. We believe our work will serve as a basis for future research dedicated to generalized, accurate, and reliable person Re-ID.

%% file: sec/supplementary.tex
\clearpage
\setcounter{page}{1}
\maketitlesupplementary

\renewcommand{\algorithmicrequire}{\textbf{Input:}}
\renewcommand{\algorithmicensure}{\textbf{Output:}}
\makeatother

\begin{algorithm}[t]
  \caption{ReMix}
  \label{alg:remix-alg}
  \begin{algorithmic}[1]
    \Require
      \Statex Encoder $\theta_{e}$,
      \Statex Momentum encoder $\theta_{m}$,
      \Statex Mini-batch size $B$,
      \Statex Number of epochs $E$,
      \Statex Number of iterations in epoch $I$,
      \Statex Labeled multi-camera data $\mathcal{D}_m$,
      \Statex Unlabeled single-camera data $\mathcal{D}_s$.
    \Ensure Trained momentum encoder $\theta_{m}$.
    
    \For{$epoch = 1$ \textit{to} $E$}
        \State Obtain embeddings $\mathcal{M}_m$ from the momentum \par encoder $\theta_{m}$ for multi-camera data $\mathcal{D}_m$;
        \State Calculate centroids and camera centroids for \par multi-camera data $\mathcal{D}_m$ using embeddings $\mathcal{M}_m$;
        \State Get pseudo labeled part $\widetilde{\mathcal{D}}_s$ of single-camera \par data $\mathcal{D}_s$, as well as embeddings $\mathcal{M}_s$ from \par the momentum encoder $\theta_{m}$ and centroids \par using \cref{alg:pseudo-labeling-alg};
        \For{$iter = 1$ \textit{to} $I$}
            \State Train $\theta_{e}$ with the general loss in \cref{eq:full-loss}: \par\hspace{0.6cm}$\mathcal{L}_{cc}$ is calculated only for $\mathcal{D}_m$, \par\hspace{0.6cm}$\mathcal{L}_{ins}$, $\mathcal{L}_{aug}$ and $\mathcal{L}_{cen}$ for $\mathcal{D}_m$ and $\widetilde{\mathcal{D}}_s$;
            \State Update $\theta_{m}$ using $\theta_{e}$ by \cref{eq:momentum};
        \EndFor
    \EndFor
  \end{algorithmic}
\end{algorithm}

\begin{algorithm}[t]
  \caption{Single-camera Data Pseudo Labeling}
  \label{alg:pseudo-labeling-alg}
  \begin{algorithmic}[1]
    \Require
        \Statex Momentum encoder $\theta_{m}$,
        \Statex Unlabeled single-camera data $\mathcal{D}_s$,
        \Statex Mini-batch size $B$,
        \Statex Number of iterations in epoch $I$.
    \Ensure pseudo labeled dataset $D$, embeddings $E$ and centroids $C$.

    \State $D \gets \emptyset$ \Comment{initialize a pseudo labeled dataset}
    \State $E \gets \emptyset$ \Comment{initialize a list of embeddings}
    \State $C \gets \emptyset$ \Comment{initialize a list of centroids}
    \State $counter \gets 0$ \Comment{pseudo labeled images counter}
    \State $limit \gets B * I$ \Comment{number of images for pseudo labeling}
    
    \While{$counter < limit$}
        \State Randomly select a video $\mathcal{V}$ from $\mathcal{D}_s$;
        \State Obtain embeddings $\widetilde{E}$ from the momentum \par encoder $\theta_{m}$ for images from the video $\mathcal{V}$;
        \State Generate a pseudo labeled dataset $\widetilde{\mathcal{D}}$ using \par embeddings $\widetilde{E}$ and DBSCAN;
        \State Calculate centroids $\widetilde{C}$ for the pseudo labeled dataset \par $\widetilde{\mathcal{D}}$ using embeddings $\widetilde{E}$;
        \State Update the pseudo labeled dataset $D$, the list of \par embeddings $E$ and the list of centroids $C$ using \par $\widetilde{\mathcal{D}}$, $\widetilde{E}$ and $\widetilde{C}$, respectively;
        \State Update $counter$ using $\widetilde{\mathcal{D}}$;
    \EndWhile
  \end{algorithmic}
\end{algorithm}

\section{Detailed Analysis}
\label{sec:detailed-analysis}

\subsection{Clustering}
\label{sec:clustering}

In ReMix, we use two types of training data --- labeled multi-camera and unlabeled single-camera data (see \cref{alg:remix-alg}). Since our method uses unlabeled single-camera data, pseudo labels are obtained for part of it at the beginning of each epoch. The pseudo labeling procedure occurs according to \cref{alg:pseudo-labeling-alg}. As we can see, our method uses DBSCAN \cite{ester1996density} for clustering, which has several parameters. One of the main parameters is the distance threshold, which regulates the maximum distance between two instances in order to consider them neighbors.

If a small distance threshold is set, then DBSCAN marks more hard positive instances as different classes. In contrast, a large distance threshold causes DBSCAN to mark more hard negative instances as the same class. Therefore, it is necessary to find the optimal value of this parameter for specific data.

\begin{figure*}[t]
    \centering
    \includegraphics[width=0.9\linewidth]{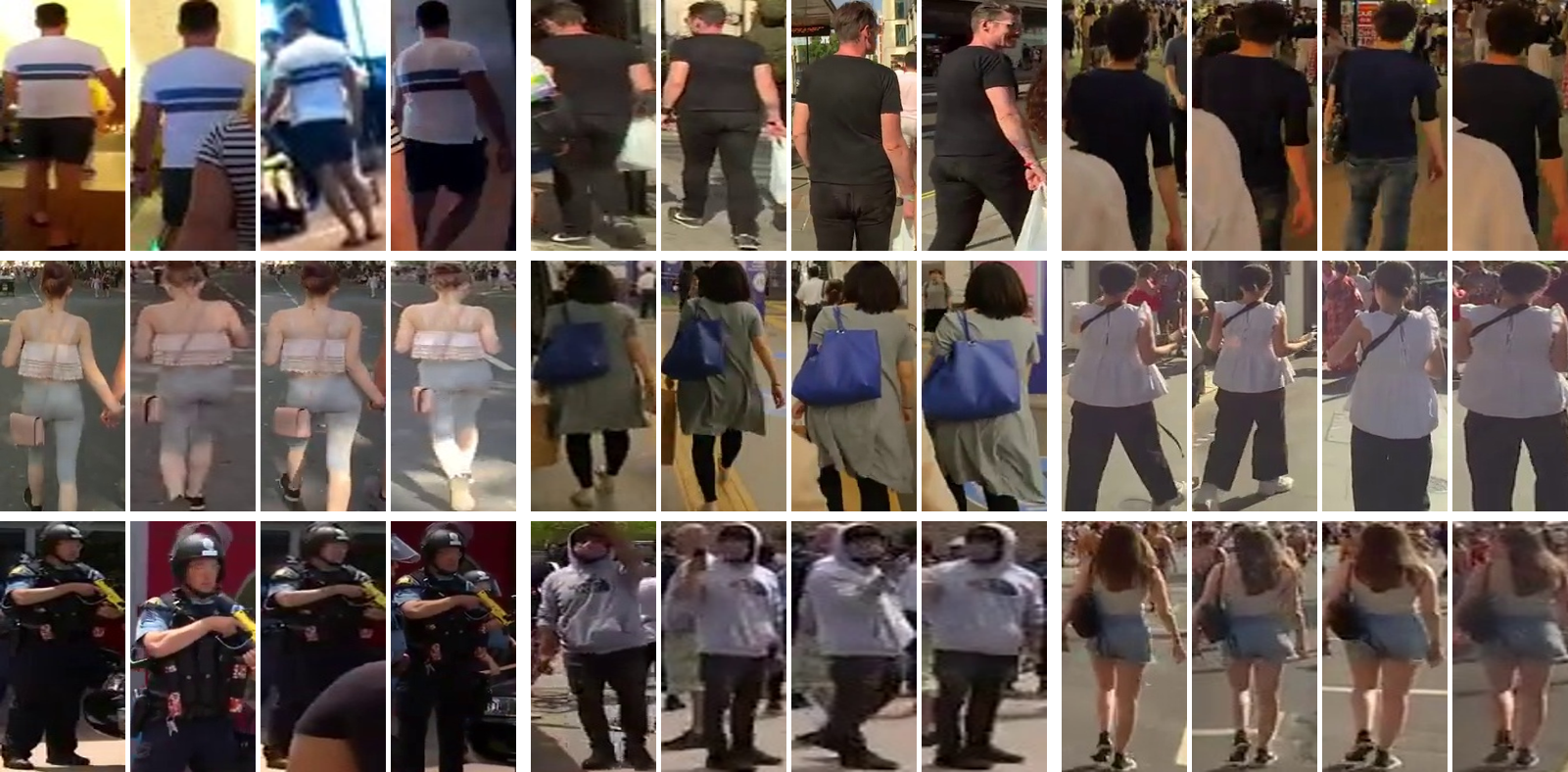}
    \caption{Examples of single-camera data clusters obtained during ReMix training. Four random images from each arbitrary cluster are selected for visualization.}
    \label{fig:clusters-examples}
\end{figure*}

In our main paper, the distance threshold is set to $0.8$, which is justified by the results of the experiments presented in \cref{tab:eps-dbscan}. Additionally, \cref{fig:clusters-examples} shows examples of single-camera data clusters obtained during ReMix training.

\begin{table}[t]
  \centering
  \begin{tabular*}{0.47\textwidth}{@{\hspace{7pt}\extracolsep{\fill}}l|*{4}{c}@{\hspace{7pt}}}
    \hline
    Threshold \hspace{1cm}& $0.65$ & $0.70$ & $0.80$ & $0.85$ \rule{0pt}{2.3ex}\rule[-0.9ex]{0pt}{0pt}\\
    \hline
    $Rank_1$ & $76.3$ & $76.3$ & $\mathbf{76.9}$ & $76.0$ \rule{0pt}{2.3ex}\\
    $mAP$ & $60.2$ & $60.5$ & $\mathbf{60.7}$ & $60.1$ \rule{0pt}{2.3ex}\\
    \hline
  \end{tabular*}
  \caption{Comparison of different distance thresholds in DBSCAN. We train the algorithm on MSMT17-merged and single-camera data from LUPerson, and test it on DukeMTMC-reID.}
  \label{tab:eps-dbscan}
\end{table}

\subsection{Mini-batch Size}
\label{sec:mini-batch-size}

In our method, we compose a mini-batch from a mixture of images from multi-camera and single-camera datasets. Let $B_m = N^m_P \times N^m_K$ be the number of images from multi-camera data in a mini-batch, and $B_s= N^s_P \times N^s_K$ be the number of images from single-camera data in a mini-batch. So, the mini-batch has a size of $B=B_m + B_s = {N^m_P \times N^m_K + N^s_P \times N^s_K}$ images (\cref{sec:data-sampling}). Here, $N^m_P$ ($N^s_P$) is the number of labels (pseudo labels) from multi-camera (single-camera) data, and $N^m_K$ ($N^s_K$) is the number of images for each label (pseudo label) from multi-camera (single-camera) data.

In our main paper, we set $N^m_P=N^s_P=8$ and $N^m_K=N^s_K=4$. Thus, the size of each mini-batch is 64 (that is, $B_m=B_s=32$ and $B=B_m + B_s=64$). We conduct several experiments to determine the impact of mini-batch size on the accuracy of ReMix. As can be seen from \cref{tab:mini-batch-sizes}, the values for parameters $B_m$ and $B_s$ selected in our main work are among the optimal ones. The experimental results given in \cref{tab:mini-batch-k-param} show a relationship between the values for parameters $N^m_K$ and $N^s_K$ and the quality of the algorithm.

Separately, it is worth noting the influence of the value for parameter $N^m_P$ on the quality of our algorithm. \cref{tab:mini-batch-sizes-multi} shows how much the accuracy of the algorithm decreases when $B_m=16$. A similar decrease in accuracy occurs with $N^m_K=8$ (see \cref{tab:mini-batch-k-param-multi}). This is because in both cases $N^m_P=4$ (in the first case, $N^m_P=B_m / N^m_K=16/4=4$; in the second case, $N^m_P=B_m / N^m_K=32/8=4$). Thus, we can conclude that the quality of ReMix is significantly affected by the number of different labels in the mini-batch.

\begin{table}[t]
  \centering
  \begin{subtable}[b]{0.495\linewidth}
    \centering
    \begin{tabular}{c|cc}
      \hline
      $B_m$ & $Rank_1$ & $mAP$ \rule{0pt}{2.3ex}\rule[-0.9ex]{0pt}{0pt} \\
      \hline
      $16$ & $69.1$ & $49.0$ \rule{0pt}{2.3ex} \\
      $32$ & $\mathbf{75.8}$ & $58.7$ \rule{0pt}{2.3ex} \\
      $64$ & $75.0$ & $\mathbf{58.9}$ \rule{0pt}{2.3ex} \\
      \hline
    \end{tabular}
    \caption{Multi-camera data.}
    \label{tab:mini-batch-sizes-multi}
  \end{subtable}
  \begin{subtable}[b]{0.495\linewidth}
    \centering
    \begin{tabular}{c|cc}
      \hline
      $B_s$ & $Rank_1$ & $mAP$ \rule{0pt}{2.3ex}\rule[-0.9ex]{0pt}{0pt} \\
      \hline
      $16$ & $77.3$ & $61.4$ \rule{0pt}{2.3ex} \\
      $32$ & $\mathbf{77.6}$ & $\mathbf{61.6}$ \rule{0pt}{2.3ex} \\
      $64$ & $77.1$ & $61.4$ \rule{0pt}{2.3ex} \\
      \hline
    \end{tabular}
    \caption{Single-camera data.}
    \label{tab:mini-batch-sizes-single}
  \end{subtable}
  \caption{Comparison of different numbers of images for each data type in a mini-batch. In "multi-camera data" experiments, we use only MSMT17-merged for training ($N^m_K=4$, $N^m_P = B_m / N^m_K$ and $B_s=0$). In "single-camera data", we train the algorithm on MSMT17-merged and single-camera data from LUPerson ($N^s_K=4$, $N^s_P = B_s / N^s_K$ and $B_m=32$). The DukeMTMC-reID dataset is used for testing in all these experiments.}
  \label{tab:mini-batch-sizes}

  \bigskip

  \centering
  \begin{subtable}[b]{0.495\linewidth}
    \centering
    \begin{tabular}{c|cc}
      \hline
      $N^m_K$ & $Rank_1$ & $mAP$ \rule{0pt}{2.3ex}\rule[-0.9ex]{0pt}{0pt} \\
      \hline
      $2$ & $\mathbf{76.0}$ & $58.5$ \rule{0pt}{2.3ex} \\
      $4$ & $75.8$ & $\mathbf{58.7}$ \rule{0pt}{2.3ex} \\
      $8$ & $70.6$ & $51.0$ \rule{0pt}{2.3ex} \\
      \hline
    \end{tabular}
    \caption{Multi-camera data.}
    \label{tab:mini-batch-k-param-multi}
  \end{subtable}
  \begin{subtable}[b]{0.495\linewidth}
    \centering
    \begin{tabular}{c|cc}
      \hline
      $N^s_K$ & $Rank_1$ & $mAP$ \rule{0pt}{2.3ex}\rule[-0.9ex]{0pt}{0pt} \\
      \hline
      $2$ & $76.6$ & $61.0$ \rule{0pt}{2.3ex} \\
      $4$ & $\mathbf{77.6}$ & $61.6$ \rule{0pt}{2.3ex} \\
      $8$ & $77.5$ & $\mathbf{62.1}$ \rule{0pt}{2.3ex} \\
      \hline
    \end{tabular}
    \caption{Single-camera data.}
    \label{tab:mini-batch-k-param-single}
  \end{subtable}
  \caption{Comparison of different values for parameters $N^m_K$ and $N^s_K$. In "multi-camera data" experiments, we use only MSMT17-merged for training ($B_m=32$, $N^m_P = B_m / N^m_K$ and $B_s=0$). In "single-camera data", we train the algorithm on MSMT17-merged and single-camera data from LUPerson ($B_s=32$, $N^s_P = B_s / N^s_K$ and $B_m=32$, $N^m_K=4$). The DukeMTMC-reID dataset is used for testing in all these experiments.}
  \label{tab:mini-batch-k-param}
\end{table}

\begin{table*}[t]
  \centering
  \begin{tabular*}{0.8\textwidth}{@{\hspace{8pt}\extracolsep{\fill}}c|c|c|cc|cc@{}}
    \hline
    \multirow{2}{*}{Image Size} & \multirow{2}{*}{Single-camera} \hspace{1pt} & \multirow{2}{*}{Inference Time*} & \multicolumn{2}{c|}{Market-1501} & \multicolumn{2}{c}{DukeMTMC-reID} \\
    & & & $Rank_1$ & $mAP$ & $Rank_1$ & $mAP$ \rule{0pt}{2.3ex}\rule[-0.9ex]{0pt}{0pt}\\
    \hline
    \multirow{2}{*}{${256 \times 128}$} &  \ding{55} & \multirow{2}{*}{90 ms} & $78.4$ & $51.7$ & $75.8$ & $58.7$ \rule{0pt}{2.3ex}\\
    & \ding{51} & & $84.0$ & $61.0$ & $77.6$ & $61.6$ \rule{0pt}{2.3ex}\\
    \hline
    \multirow{2}{*}{${384 \times 128}$} & \ding{55} & \multirow{2}{*}{149 ms} & $79.2$ & $51.3$ & $76.2$ & $59.3$ \rule{0pt}{2.3ex}\\
    & \ding{51} & & $\mathbf{85.1}$ & $\mathbf{62.7}$ & $\mathbf{78.4}$ & $\mathbf{63.3}$ \rule{0pt}{2.3ex}\\
    \hline
  \end{tabular*}
  \begin{tablenotes}\small
      \centering
      \item[1] * Inference speed is estimated in a single-core test on the Intel Core i7-9700K.
  \end{tablenotes}
  \caption{Comparison of different input image sizes. We train the algorithm on MSMT17-merged and single-camera data from LUPerson (where applicable), and test it on DukeMTMC-reID.}
  \label{tab:imgs-sizes}
  
  \bigskip

  \centering
  \begin{tabular*}{0.8\textwidth}{@{\hspace{8pt}\extracolsep{\fill}}c|c|c|cc|cc@{}}
    \hline
    \multirow{2}{*}{Architecture} & \multirow{2}{*}{Single-camera} \hspace{0.5pt} & \multirow{2}{*}{Inference Time*} & \multicolumn{2}{c|}{Market-1501} & \multicolumn{2}{c}{DukeMTMC-reID} \\
    & & & $Rank_1$ & $mAP$ & $Rank_1$ & $mAP$ \rule{0pt}{2.3ex}\rule[-0.9ex]{0pt}{0pt}\\
    \hline
    \multirow{2}{*}{ResNet50-IBN} &  \ding{55} & \multirow{2}{*}{90 ms} & $78.4$ & $51.7$ & $75.8$ & $58.7$ \rule{0pt}{2.3ex}\\
    & \ding{51} & & $\mathbf{84.0}$ & $\mathbf{61.0}$ & $\mathbf{77.6}$ & $\mathbf{61.6}$ \rule{0pt}{2.3ex}\\
    \hline
    \multirow{2}{*}{ResNet50} & \ding{55} & \multirow{2}{*}{82 ms} & $76.0$ & $46.8$ & $72.4$ & $53.5$ \rule{0pt}{2.3ex}\\
    & \ding{51} & & $78.4$ & $53.8$ & $73.6$ & $56.4$ \rule{0pt}{2.3ex}\\
    \hline
  \end{tabular*}
  \begin{tablenotes}\small
      \centering
      \item[1] * Inference speed is estimated in a single-core test on the Intel Core i7-9700K.
  \end{tablenotes}
  \caption{Comparison of different encoder architectures. We train the algorithm on MSMT17-merged and single-camera data from LUPerson (where applicable), and test it on DukeMTMC-reID.}
  \label{tab:archs}
\end{table*}

\subsection{Input Image Size}
\label{sec:input-images-size}

Most works devoted to the person re-identification task use input images of size ${256 \times 128}$ pixels. Input images of the same size are used in our method. However, after studying other state-of-the-art methods in detail, we noticed that \cite{liao2020interpretable, liao2022graph, liao2021transmatcher} use larger input images --- ${384 \times 128}$ pixels.

We conducted several experiments to analyze the quality of ReMix with this size of the input images. The results of these experiments are shown in \cref{tab:imgs-sizes}. As can be seen, the accuracy of our method improves as the size of the input images increases. It is worth noting that the joint use of labeled multi-camera and unlabeled single-camera data for training also has a beneficial effect on the quality of Re-ID with larger input images. This further confirms the effectiveness of the proposed ReMix method.

Obviously, the use of larger input images can significantly increase the computational costs of the algorithm. This is confirmed by the estimates given in \cref{tab:imgs-sizes}. Therefore, in our main work, we choose to prioritize method performance and resize all input images to ${256 \times 128}$.

Separately, we note that according to \cref{tab:comparison}, ReMix using ${256 \times 128}$ input images outperforms others (including those methods that use ${384 \times 128}$ input images) in the cross-dataset scenario. Thus, our method achieves high accuracy while also being computationally efficient, which is important for practical applications.

\subsection{Encoder Architecture}
\label{sec:encoder-arch}

In \cite{pan2018two, jia2019frustratingly, zhou2021learning} it was shown that using combinations of Batch Normalization and Instance Normalization improves the generalization ability of neural networks. Therefore, we compare two encoder architectures in ReMix: ResNet50 \cite{he2016deep} and ResNet50-IBN (ResNet50 with IBN-a layers) \cite{pan2018two}. ResNet50-IBN differs from ResNet50 only in that the former uses Instance Normalization in addition to Batch Normalization. The results of our comparison presented in \cref{tab:archs} also demonstrate the effectiveness of ResNet50 with IBN-a layers in the cross-dataset scenario.

Moreover, our experiments show that joint training on a mixture of multi-camera and single-camera data significantly improves the accuracy of the algorithm, even when ResNet50 is used as the encoder and the momentum encoder. Additionally, according to the speed estimation of our algorithm with different encoder architectures, ResNet50-IBN is slower than ResNet50 by less than 10 ms. Therefore, the use of ResNet50 with IBN-a layers in our main paper is justified, as this architecture represents a trade-off between quality and speed.

\begin{table*}[t]
  \centering
  \begin{tabular}{@{\hspace{5pt}}l|c|cc|cc|cc@{\hspace{3pt}}}
    \hline
    \multirow{2}{*}{Method} & \multirow{2}{*}{Reference} & \multicolumn{2}{c|}{Market-1501} & \multicolumn{2}{c|}{DukeMTMC-reID} & \multicolumn{2}{c}{MSMT17}\\
    & & $Rank_1$ & $mAP$ & $Rank_1$ & $mAP$ & $Rank_1$ & $mAP$\rule{0pt}{2.3ex}\rule[-1.0ex]{0pt}{0pt}\\
    \hline
    ISP \cite{zhu2020identity} & ECCV20 & $94.2$ & $84.9$ & $86.9$ & $75.6$ & --- & ---\rule{0pt}{2.3ex}\\
    RGA-SC \cite{zhang2020relation} & CVPR20 & $\underline{96.1}$ & $88.4$ & --- & --- & $80.3$ & $57.5$\rule{0pt}{2.3ex}\\
    FlipReID \cite{ni2021flipreid} & EUVIP21 & $95.3$ & $88.5$ & $89.4$ & $79.8$ & $83.3$ & $64.3$\rule{0pt}{2.3ex}\\
    CAL \cite{rao2021counterfactual} & ICCV21 & $94.5$ & $87.0$ & $87.2$ & $76.4$ & $79.5$ & $56.2$\rule{0pt}{2.3ex}\\
    CDNet \cite{li2021combined} & CVPR21 & $95.1$ & $86.0$ & $88.6$ & $76.8$ & $78.9$ & $54.7$\rule{0pt}{2.3ex}\\
    LTReID \cite{wang2022ltreid} & TMM22 & $95.9$ & $89.0$ & $\underline{90.5}$ & $80.4$ & $81.0$ & $58.6$\rule{0pt}{2.3ex}\\
    DRL-Net \cite{jia2022learning} & TMM22 & $94.7$ & $86.9$ & $88.1$ & $76.6$ & $78.4$ & $55.3$\rule{0pt}{2.3ex}\\
    Nformer \cite{wang2022nformer} & CVPR22 & $94.7$ & $\underline{91.1}$ & $89.4$ & $\mathbf{83.5}$ & $77.3$ & $59.8$\rule{0pt}{2.3ex}\\
    CLIP-ReID \cite{li2023clip} & AAAI23 & $95.7$ & $89.8$ & $90.0$ & $80.7$ & $84.4$ & $63.0$\rule{0pt}{2.3ex}\\
    AdaSP \cite{zhou2023adaptive} & CVPR23 & $95.1$ & $89.0$ & $\mathbf{90.6}$ & $\underline{81.5}$ & $84.3$ & $\underline{64.7}$\rule{0pt}{2.3ex}\\
    SOLIDER* \cite{chen2023beyond} & CVPR23 & $\underline{96.1}$ & $\mathbf{91.6}$ & --- & --- & $\mathbf{85.9}$ & $\mathbf{67.4}$\rule{0pt}{2.3ex}\\
    ReMix (w/o s-cam.) & Ours & $94.7$ & $84.9$ & $87.9$ & $75.8$ & $83.9$ & $62.8$\rule{0pt}{2.3ex}\\
    ReMix & Ours & $\mathbf{96.2}$ & $89.8$ & $89.6$ & $79.8$ & $\underline{84.8}$ & $63.9$\rule[-1.0ex]{0pt}{0pt}\\
    \hline
  \end{tabular}
  \begin{tablenotes}\small
      \centering
      \item[1] * This is a transformer-based method.
  \end{tablenotes}
  \caption{Comparison of our ReMix method with others in the standard person Re-ID task. In this comparison, we use two versions of the
proposed method: without using single-camera data and with using single-camera data during training. Here, we use the LUPerson dataset
as single-camera data to train ReMix. In this table, bold and underlining fonts suggest the best and the second-best performance, respectively.}
  \label{tab:comparison-standard}
\end{table*}

\begin{figure*}[t]
    \centering
    \includegraphics[width=0.88\linewidth]{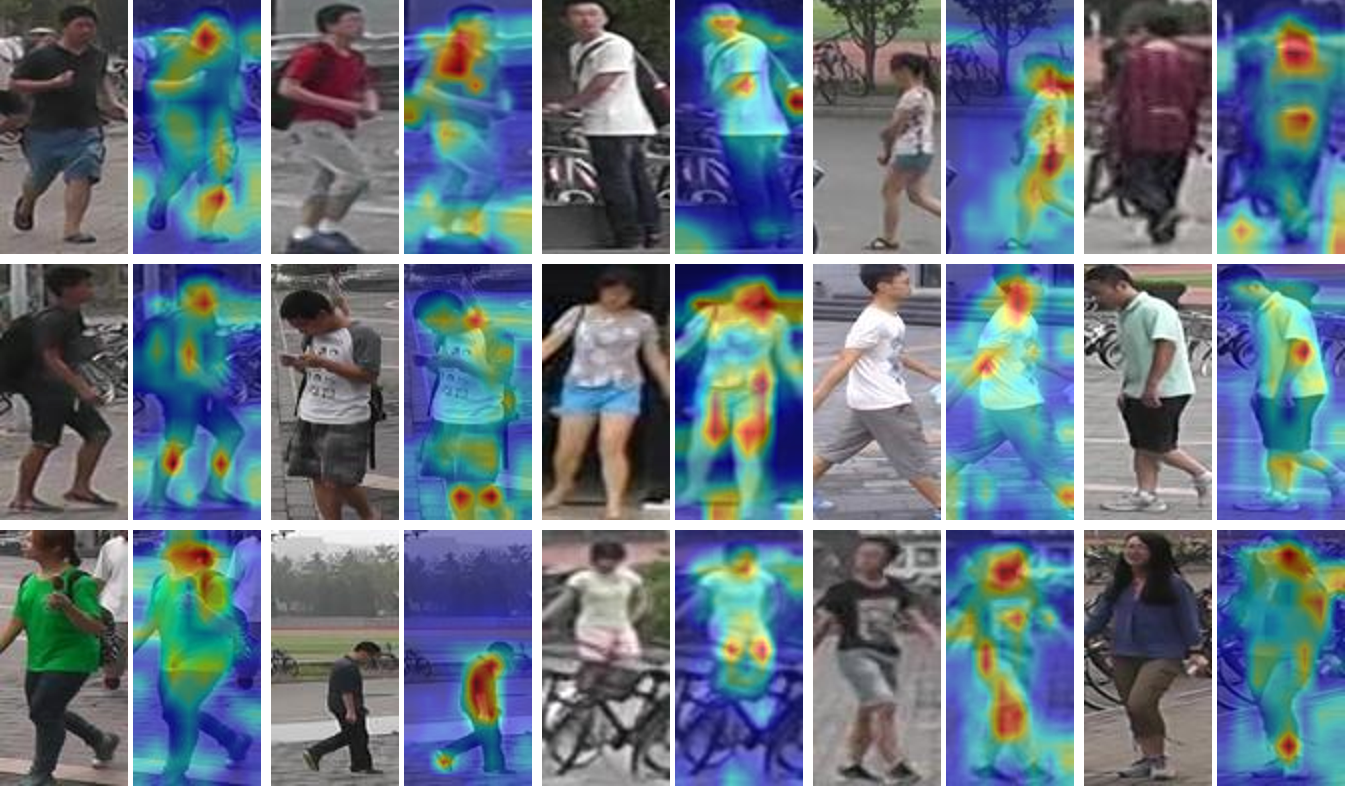}
    \caption{Visualization of activation maps of ReMix on the Market-1501 dataset.}
    \label{fig:activations-visualization}
\end{figure*}

\section{Standard Person Re-ID}

In our main paper, we aim to improve the generalization ability of person Re-ID methods. Our experiments in the cross-dataset and multi-source cross-dataset scenarios show that our ReMix method has a high generalization ability and outperforms state-of-the-art methods in the generalizable person Re-ID task (\cref{sec:comparison}). We choose these test protocols because they are the closest to real-world applications of Re-ID algorithms. Indeed, in real-world scenarios, we do not have prior information about the features of capturing environments in an arbitrary scene. Therefore, person Re-ID methods should have a high generalization ability and work with acceptable accuracy in almost all possible scenes.

Even so, as we can see from \cref{tab:comparison-standard}, our method shows competitive accuracy in the standard person Re-ID task (when trained and tested on separate splits of the same dataset). It is worth noting that the other methods in this comparison are designed specifically for standard person Re-ID scenario. At the same time, ReMix is intended as a method with high generalization ability, which should perform well in various scenes. In other words, our ReMix method is not adapted to work with a specific scene, unlike competitors. Thus, such a strong performance in this task clearly indicates the consistency and flexibility of ReMix, as well as the effectiveness of using single-camera data in addition to multi-camera data during training.

\section{Tracking}

\begin{table}[t]
  \centering
  \begin{tabular*}{0.47\textwidth}{@{\hspace{7pt}}c|c|cc|cc@{\hspace{3pt}}}
    \hline
    \multirow{2}{*}{Hz} & \multirow{2}{*}{S-cam.} & \multicolumn{2}{c|}{MOT15} & \multicolumn{2}{c}{MOT17}\\
    & & $MOTA$ & $IDsw$ & $MOTA$ & $IDsw$\\
    \hline
    \multirow{2}{*}{$2$} & \ding{55} & $83.8$ & $70$ & $73.8$ & $249$\rule{0pt}{2.3ex}\\
    & \ding{51} & $\mathbf{84.6}$ & $\mathbf{66}$ & $\mathbf{76.9}$ & $\mathbf{219}$\rule[-1.0ex]{0pt}{0pt}\\
    \hline
    \multirow{2}{*}{$4$} & \ding{55} & $85.8$ & $105$ & $80.5$ & $333$\rule{0pt}{2.3ex}\\
    & \ding{51} & $\mathbf{88.0}$ & $\mathbf{90}$ & $\mathbf{83.1}$ & $\mathbf{288}$\rule[-1.0ex]{0pt}{0pt}\\
    \hline
    \multirow{2}{*}{$8$} & \ding{55} & $91.6$ & $120$ & $88.6$ & $375$\rule{0pt}{2.3ex}\\
    & \ding{51} & $\mathbf{93.2}$ & $\mathbf{99}$ & $\mathbf{90.6}$ & $\mathbf{308}$\rule[-1.0ex]{0pt}{0pt}\\
    \hline
  \end{tabular*}
  \caption{Impact of using single-camera data in ReMix in the tracking task. In these experiments, we use MSMT17-merged and single-camera data from LUPerson (where applicable) for ReMix training. The Deep SORT algorithm is used as a tracking method.}
  \label{tab:ablation-tracking}
\end{table}

Re-ID methods are often used as components of more practical applications, such as tracking. For example, in Deep SORT \cite{Wojke2017simple}, the Re-ID algorithm is used to bind detections from different frames into tracks. We conduct experiments to study the impact of using single-camera data in addition to multi-camera data in ReMix not only on the quality of person Re-ID, but also on tracking.

In this study, we apply our implementation of the Deep SORT algorithm as a tracking method, using two versions of the proposed Re-ID method: without using single-camera data and with using single-camera data during training. We employ the training parts of the MOT15 \cite{leal2015motchallenge} and MOT17 \cite{milan2016mot16} benchmarks as the tracking test datasets (important: these datasets are not used to train ReMix). Since the tracking quality depends on many factors (e.g., the object detector), we use public detections from MOT15 and MOT17 to demonstrate the effectiveness of our Re-ID algorithm. In our experiments, we use Multi-Object Tracking Accuracy ($MOTA$) \cite{bernardin2008evaluating} and Number of Identity Switches ($IDsw$) \cite{li2009learning} metrics to evaluate tracking performance. Additionally, to demonstrate the effectiveness of ReMix for binding detections from different frames into tracks, we test Deep SORT with different frame rates: 2, 4, and 8 Hz.

As can be seen from \cref{tab:ablation-tracking}, the use of single-camera data in addition to multi-camera data in ReMix has a beneficial effect not only on the quality of person Re-ID, but also on tracking. With different frame rates on both benchmarks, the tracking algorithm with the proposed Re-ID method using single-camera data during training performs best. This further demonstrates the effectiveness and flexibility of ReMix. It is also important to note that in this study, we do not aim to achieve state-of-the-art results in the tracking task, but rather to demonstrate the effectiveness of our Re-ID method.